\pdfoutput=1

\documentclass[11pt]{article}

\usepackage[]{acl}

\usepackage{times}
\usepackage{latexsym}
\usepackage{multirow}
\usepackage{kantlipsum}
\usepackage{algorithm}
\usepackage{algpseudocode}
\usepackage{times}
\usepackage{latexsym}

\usepackage{graphicx}
\usepackage{lipsum}
\usepackage{tabularx}
\usepackage{enumitem}
\newlist{steps}{enumerate}{1}
\setlist[steps, 1]{label = Step \arabic*:,  align=left, leftmargin=*]}
\usepackage[T1]{fontenc}

\usepackage[utf8]{inputenc}

\usepackage{microtype}

\newcommand*{\affaddr}[1]{#1} 
\newcommand*{\affmark}[1][*]{\textsuperscript{#1}}
\newcommand*{\email}[1]{\texttt{#1}}
\title{Knowledge Graph Construction and Its Application in Automatic Radiology Report Generation from Radiologist's Dictation}

\author{%
\textbf{Kaveri Kale}\affmark[1], \textbf{Pushpak Bhattacharyya}\affmark[1], \\
\textbf{Aditya Shetty}\affmark[2], \textbf{Milind Gune}\affmark[2], \textbf{Kush Shrivastava}\affmark[3], \textbf{Rustom Lawyer}\affmark[3] and \textbf{Spriha Biswas}\affmark[3]\\
\affaddr{\affmark[1]Indian Institute of Technology Bombay, India},
\affaddr{\affmark[2]Consultant Radiologist, India}, 
\affaddr{\affmark[3]Augnito India Pvt Ltd, India}\\
\email{\{kaverikale, pb\}@cse.iitb.ac.in}, \email{adityashetty01@gmail.com}, \\
\email{dgune@rediffmail.com}, \email{\{kush.shrivastava, rustom, spriha.biswas\}@augnito.ai}\\
}

\begin{document}
\maketitle
\begin{abstract}
Conventionally, the radiologist prepares the diagnosis notes and shares them with the transcriptionist. Then the transcriptionist prepares a preliminary formatted report referring to the notes, and finally, the radiologist reviews the report, corrects the errors, and signs off. This workflow causes significant delays and errors in the report. In current research work, we focus on applications of NLP techniques like Information Extraction (IE) and domain-specific Knowledge Graph (KG) to automatically generate radiology reports from radiologists’ dictation. This paper focuses on KG construction for each organ by extracting information from an existing large corpus of free-text radiology reports. We develop an information extraction pipeline that combines rule-based, pattern-based, and dictionary-based techniques with lexical-semantic features to extract entities and relations. Missing information in short dictation can be accessed from the KGs to generate pathological descriptions and hence the radiology report. Generated pathological descriptions evaluated using semantic similarity metrics, which shows 97\% similarity with gold standard pathological descriptions. Also, our analysis shows that our IE module is performing better than the OpenIE tool for the radiology domain. Furthermore, we include a manual qualitative analysis from radiologists, which shows that 80-85\% of the generated reports are correctly written, and the remaining are partially correct.
\end{abstract}

\section{Introduction}
\label{sec:introduction}
Radiology is an integral part of medical care. Radiological imaging-based evidence (X-ray, MRI, CT, USG, \textit{etc.}) is crucial in determining the nature of the treatment. The usual radiology process is as follows: A patient gets scanned. Then the radiologist prepares the diagnosis notes (referred to as \textbf{\small{radiologist’s dictation}}) by either dictating to a voice recording device or writing it on paper and hands over the notes to a transcriptionist. The transcriptionist opens a scan-specific standardized template corresponding to all normal findings (referred to as \textbf{\small{normal report template}}) and edits it based on the measurements and findings reported by the radiologist in more descriptive form (referred to as \textbf{\small{pathological description}}).
\subsection{Motivation}
Radiologists are in big demand since the ratio of radiologists per number of patients is very low. Ratio in India is, 1:100,000, the corresponding ratio in the US is 1:10,000, and for China, it is 1:14,772 \citep{arora2014training}. It results in very high patient inflows, making radiologists incredibly busy and stressed out. Currently adopted workflow causes (i) significant delays in report turnaround time, (ii) errors in the reports, and (iii) burnout. These challenges amplify further, because in densely populated countries, radiologists handle several patients every day. Our interactions with radiologists, diagnostic centers, and hospitals highlight that many radiologists want to eliminate the tedious report typing process and focus on the diagnosis. 

To overcome tedious typing, radiologists resort to dictating observations in short or abstract form. Domain specific knowledge is required to generate the correct pathological description from short dictations. Once we generate patient-specific pathological description from radiologist's dictation then we can add it in normal report template by replacing corresponding normal sentences. 

Domain knowledge can be acquired from already existing radiology free-text reports. We need a structured format of all essential medical information to reuse it. There are different technologies used to store the structured data like, relational databases, xml files, KGs, \textit{etc.} KGs are used solely for deriving insights. Maintaining a KG is worry-free because we don’t have to think about how the additional data stored in the graph will affect the existing data. Using a KG to uncover insights is a better choice when the system needs domain insight.

\subsection{Problem Statement}Design a system that generates a structured patient-specific report from radiologist's dictation and domain knowledge.
\begin{itemize}
    \item Input: (i) Radiologist's dictation, and (ii) Normal report template.
    \item Output: Radiology report with patient-specific findings.

\end{itemize}
Domain knowledge will come from KG.\\

\textbf{Sub-Problem: }Develop a system that automatically constructs a KG of essential medical information in radiology free-text reports. 
\begin{itemize}
    \item Input: Radiology free-text report corpus.
    \item Output: Structured representation of the essential medical information contained within the free-text reports in a hierarchical KG.
\end{itemize}

\subsection{Challenges} 
Radiologists help us to construct a preliminary KG of the skeleton (higher-level hierarchy) for the anatomy and findings of an organ. Manually identifying all the entities in the radiology domain is also tricky for radiologists. Hence to complete the preliminary KG provided by the radiologists, we extract information from the radiology report corpus. 

IE is one of the foremost steps in KG construction \citep{zhao2018architecture}. Information extraction from free-text clinical notes/narratives, such as radiology reports, is difficult due to the nuances of natural language like misspellings, abbreviations, and non-standard terminologies. There are three main approaches for IE: (i) Machine Learning (ML) based technique, (ii) Rule-based technique, (iii) Pattern-based technique, and iv) Lexicon dictionary-based technique. The ML-based technique needs annotated corpus. We do not have annotated data, and annotating data in-house is time-consuming and costly. Hence we are not considering ML-based techniques for IE. Pattern-based IE uses a lexico-syntactic and semantic pattern dictionary to extract information from free-text, as detailed in the paper \citep{tang2008information}. The construction of a pattern dictionary is required in a pattern-based approach. To extract information from text, the rule-based technique employs a set of general rules. The rule-based and pattern-based techniques need domain understanding and language understanding. The lexicon dictionary-based approach extracts entities present in the dictionary, but it fails to extract entities not present in the dictionary. Hence, the only pattern-based, rule-based, or lexicon dictionary-based approach does not fit to the radiology report domain.

We propose an approach for information extraction that combines rule-based, pattern-based, and lexicon dictionary-based techniques with lexical semantic features.  We test the efficacy of constructed KGs by calculating evaluation metrics BLEU \citep{papineni2002bleu} score and ROUGE \citep{lin2004rouge} score of pathological description generated using constructed KG vs. gold standard pathological descriptions generated by radiologists manually. We also evaluate our IE system by calculating precision, recall, and F-score of system extracted triples vs. gold standard triples and compare it with the OpenIE system.
\subsection{Our Contributions}

The key contributions are as follows:
\begin{enumerate}
    \item Preliminary KGs construction by radiologist manually.
    \item Information extraction pipeline to extract medical entities and relations from radiology reports. It extracts all necessary entities from the radiology text reports (findings, observations, anatomy, modifiers, and properties) and their relations.
    \item KG augmentation using extracted triples from reports and store constructed KG in standard Resource Description Framework (RDF) triple format.\footnote{There are no KGs constructed before for all body organs, including all necessary information like findings, observations, anatomy, properties, and modifiers related to the organ. In our work, we construct KGs for the Liver, Kidney, Gallbladder, Uterus, Urinary bladder, Ovary, Pancreas, Prostate, Biliary Tree, and Bowel, \textit{etc.} for Ultrasound scan procedure. We have developed generic pipeline for radiology domain to construct KGs that can be used for CT, MRI, X-ray, \textit{etc.} scan procedures.}
    \item Pathological description generation from radiologist's dictation using constructed KGs. 
    \item Map generated pathological description in normal report template at the appropriate location to generate patient-specific report.
 
\end{enumerate}

\section{Related Work}
Research is done in automatic radiology report generation based on scanned images. \cite{yuan2019automatic} proposes an automated structured-radiology report generation system using extracted features from images. \cite{loveymi2021automatic} proposed a system that generate descriptions for natural images by image captioning.

There is a wealth of research done on building medical KG from Electronic Medical Records (EMR). \cite{finlayson2014building} builds a graph from medical text, clinical notes \textit{etc.} Graph nodes represents diseases, drugs, procedures, and devices. \cite{rotmensch2017learning} uses the EMR to construct the graph of diseases and symptoms. Researchers worked on creating medical KG from EMR, but no one has built a KG for the radiology domain except \cite{zhang2020radiology}. Graph embedding module is proposed by \cite{zhang2020radiology} that helps to generate radiology reports from image reports. Each node in their KG represents disease.
\cite{taira2001automatic} developed an NLP pipeline to structure the critical medical information. Extracted information includes the existence, location, properties, and diagnostic interpretation of findings from radiology free-text documents. Information is not integrated since they store the structured information for each report separately. Also, this system does not accept reports with different reporting styles. However, this is not always the case. Every radiologist has their dictation style and reporting style. 

IE systems that are based on IE patterns are surveyed by \cite{muslea1999extraction}. \cite{ghoulam2015information} extracts signs of lung cancer, anatomical location, and relation between the signs and the locations expressed in the radiology reports. \cite{embarek2008learning} used a morpho-syntactic patterns in their rule-based method to find medical entities like symptoms, disease, exams, medicament, and treatment. \cite{xu2009discovery} explains that pattern is a sub dependency tree that indicates a relation instance. \cite{pons2016natural} gives the overview of NLP techniques that can be used in radiology.

\section{Our Approach}

\begin{figure}[!t]
\centerline{\includegraphics[width=0.9\columnwidth]{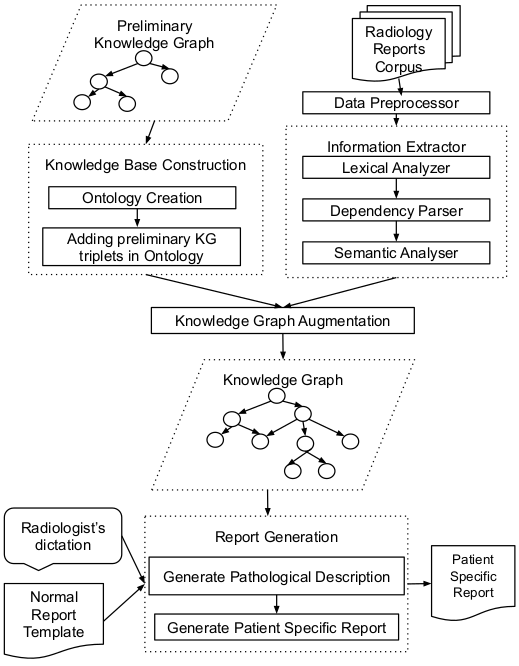}}
\caption{The architecture of our system. KG construction and patient-specific report generation are the two main modules in our system.}
\label{fig:sys_design}
\end{figure}
\autoref{fig:sys_design} shows the system architecture. Our system converts the radiologist’s dictation to a patient-specific report. The main tasks of the system are: 
\begin{enumerate}
    \item construct organ wise KGs using preliminary KG provided by radiologists and radiology report corpus
    \item generate pathological description using KGs
    \item map generated pathological descriptions to the corresponding location in the normal report template
    \item eliminate normal sentences from templates where an abnormal finding has been reported
    \item output the patient-specific report
\end{enumerate}

\subsection{Knowledge Graph Construction}
Paulheim and Heiko \citep{paulheim2017knowledge} defines Knowledge Graph (KG) as "A knowledge graph (i) mainly describes real-world entities and their interrelations, organized in a graph, (ii) defines possible classes and relations of entities in a schema, (iii) allows for potentially interrelating arbitrary entities with each other and (iv) covers various topical domains." KGs are designed with suitable ontology to store domain knowledge. Domain ontology and individual information together form a Knowledge Base (KB). 

KGs are constructed based on anonymised radiology reports\footnote{Anonymised radiology reports were used as provided by a company collaborating with us, with due consent of the physicians.}. The de-identified corpus consists of approximately a hundred thousand reports. These include CT, MRI, Ultrasound, and X-Ray reports. Since reports are collected from different hospitals, these report's reporting styles and dictation styles are different. 
\\
\subsubsection{Knowledge Base Creation}
\paragraph{\textbf{Ontology Creation}}
Ontologies are semantic data models that define the types of things in a specific domain and the properties used to describe those types. Three main components of ontology are \textbf{\small{Classes}}, \textbf{\small{Relationships}}, and \textbf{\small{Attributes}}.

\begin{figure}[h]
\centering
\includegraphics[width=0.7\columnwidth]{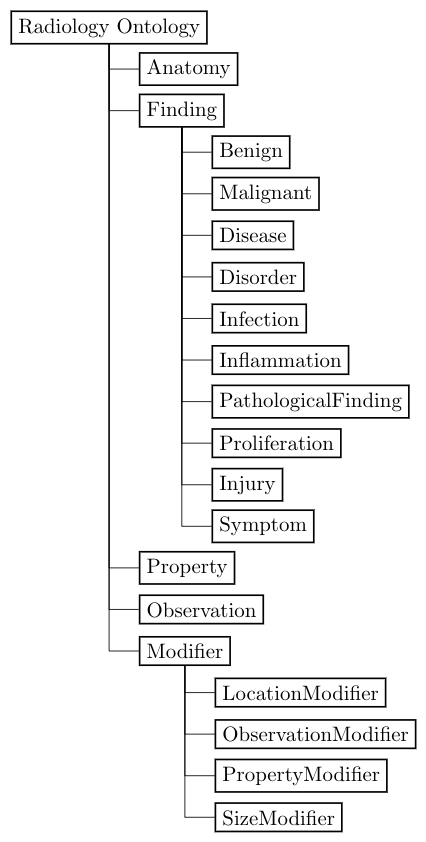}
\caption{The class hierarchy of radiology ontology that we have created. Anatomy, Finding, Observation, Property, and Modifier are the higher-level classes in the hierarchy.}
\label{fig:rado}
\end{figure}

We have created our ontology by integrating several classes of Radlex\footnote{\url{http://radlex.org/} \\RadLex is licensed freely for commercial and non-commercial users.} entities. For example, Radlex descriptor subclasses are combined in \textbf{Modifier} class, several subclasses of Radlex observation into \textbf{Observation} class, several subclasses of Radlex anatomical entity into \textbf{Anatomy} class, \textit{etc}. We believe that leaving the granularity at a coarse level is a practical choice to make ontology more generic. We have defined eight logical relations \textbf{\small{PartOf}}, \textbf{\small{TypeOf}}, \textbf{\small{ModifierOf}}, \textbf{\small{ObservationOf}}, \textbf{\small{DefaultObservationOf}}, \textbf{\small{PropertyOf}}, \textbf{\small{DefaultPropertyOf}} and \textbf{\small{FoundIn}}. We kept our ontology very simple. However, it covers all entities in the radiology report domain. We define attributes like \textbf{\small{preferedName}}, \textbf{\small{synonyms}} and \textbf{\small{wordForms}} for all classes in our ontology. \autoref{fig:rado} indicates higher level class hierarchy of the ontology that we have constructed with the help of domain experts. Protégé\footnote{\url{https://protege.stanford.edu/}} \citep{musen2015protege} suite is used for ontology development.

The purpose of constructing KGs is that we should get all the necessary information which is missing in abstract text from radiology. For example, we should get the anatomical location of finding from KG if it is not mentioned in the text. \textit{\color{darkgray}The lesion is found in segment VI of the liver}. Here anatomical location is \textit{\color{darkgray}segment VI} of the \textit{\color{darkgray}right lobe} of the \textit{\color{darkgray}liver}. The \textit{\color{darkgray}right lobe} is missing in the text, and the KG gives the missing location. Another example is that if some finding is mentioned without details, we should get all default observations and properties from KG. For example, \textit{\color{darkgray}Acute pancreatitis}, its default observations and properties are the presence of \textit{\color{darkgray}peripancreatic fluid}, \textit{\color{darkgray}increased size} of the \textit{\color{darkgray}pancreas}, \textit{\color{darkgray}inhomogeneous echotexture}, \textit{etc.} 

We have constructed KGs in hierarchical structures. Root represents the organ, and their children express their anatomy (i.e., parts). The edge between them is labeled by \textbf{\small{PartOf}} relation. Similarly, if any particular radiological observation is found for an organ or its parts, that observation can be added as a child of that anatomical entity. Furthermore, we label the edge between them by \textbf{\small{\color{darkgray}FoundIn}} relation.
\\
\paragraph{\textbf{Preliminary Knowledge Graph}}
\begin{figure*}[h]
\includegraphics[width=\textwidth]{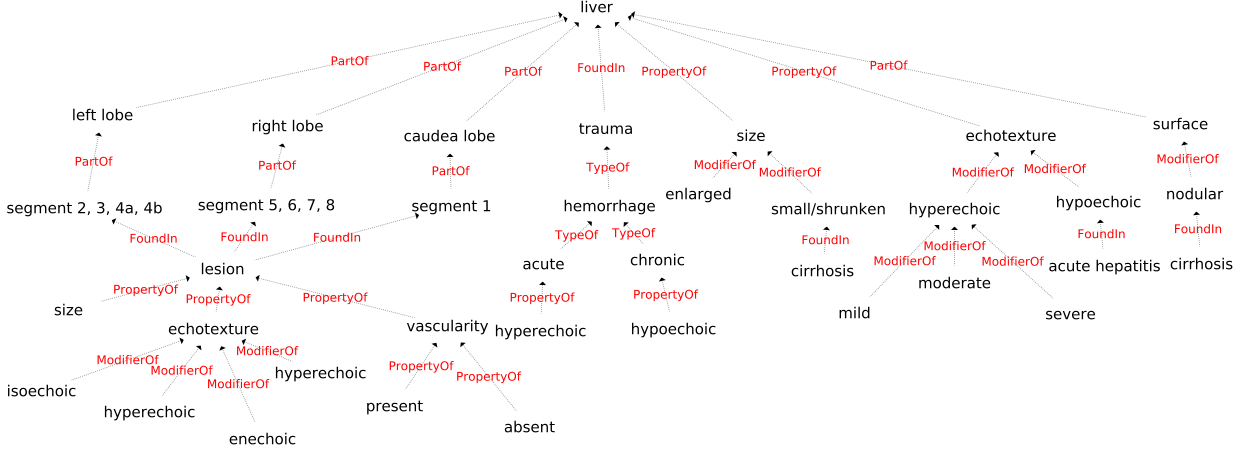}
\caption{Preliminary KG of Liver that is constructed by domain experts manually. Preliminary KGs only give higher-level anatomy, observations, findings, and properties. The preliminary KG does not include concrete or deeper level entities and relations.}
\label{liverkg}
\end{figure*}
 Domain experts provide preliminary KGs for each organ with higher-level anatomy and pathology corresponding to that organ. We convert preliminary KGs into standard RDF \citep{miller1998introduction} triples format. Using transformation rules\footnote{\url{https://github.com/protegeproject/cellfie-plugin}} we have loaded preliminary KG triplets as individuals/instances in Protege’ tool.
\autoref{liverkg} shows the KG of the Liver, which is provided by radiologists.

\subsubsection{Data Preprocessor}

To construct KGs, we use a radiology report corpus. The data preprocessor module takes radiology reports as input. Radiology reports contain Header, Pathology Description, History, and Conclusion/Impression sections. We use simple heuristics like regular expressions to fetch only the Pathological Description and Impression section. We use spell corrector and word-tokenization algorithms since the extracted sentences contain spelling mistakes, unwanted punctuation marks, \textit{etc.} Spelling correction and word-segmentation algorithm implemented using Symspell\footnote{\url{https://github.com/wolfgarbe/symspell}} APIs. We use the symmetric delete spelling correction algorithm, which provides much higher speed and lower memory consumption. We apply the domain dictionary of correct words and the frequency of their occurrences in the corpus to correct the spellings.
\\
\subsubsection{Information Extraction}
The information extraction process extracts the entities from unstructured text and couples those entities with their relations. We use the same logical relations that we have defined in our ontology.
For example, the corpus sentence is \textit{\color{darkgray}Right kidney is normal in size, shape, location and cortical echogenicity.} Here, \textit{\color{darkgray}size, shape, location} and \textit{\color{darkgray}echogenicity} have a relation with the \textit{\color{darkgray}right kidney}. \textit{\color{darkgray}Cortical} and \textit{\color{darkgray}echogenicity} are related to each other by relation \textbf{\small{ModifierOf}}. Also \textit{\color{darkgray}normal} is a modifier of \textit{\color{darkgray}size, shape, location}, and \textit{\color{darkgray}echogenicity}. However, sate-of-the-art OIE tools like Stanford OpenIE\footnote{\url{https://nlp.stanford.edu/software/openie.html}} are not capable of extracting these concrete relations from free-text \citep{etzioni2008open}.
As shown in \autoref{tbl:openie}, for the sentence 1, OpenIE could not find the relation between \textit{\color{darkgray}calculus} and \textit{\color{darkgray}middle calyx}, \textit{\color{darkgray}middle calyx} and \textit{\color{darkgray}right kidney} and for the sentence 2, OpenIE does not consider the \textit{\color{darkgray}shape}, \textit{\color{darkgray}location}, and \textit{\color{darkgray}cortical echogenicity}.

\begin{table}[htb]
\resizebox{\columnwidth}{!}{%
\begin{tabular}{|p{0.38\columnwidth}|p{0.62\columnwidth}|} 
\hline
\textbf{Input Sentence} & \textbf{Triples Extracted Using OpenIE}
 \\
\hline

\footnotesize{A 5 mm calculus is noted in an upper calyx and a 4 x 3 mm calculus is noted in a middle calyx of right kidney.} & 
\begin{tabular}[t]{l}\footnotesize{(mm calculus, is noted in, upper calyx)}\\
\footnotesize{(mm calculus, is, noted)}\\
\footnotesize{(5 mm calculus, is noted in, calyx)}\\
\footnotesize{(5 mm calculus, is noted in, upper calyx)}\\
\footnotesize{(5 mm calculus, is, noted)}\\
\footnotesize{(mm calculus, is noted in, calyx)}

\end{tabular}
\\
\hline

\footnotesize{Right kidney is normal in size 9.6 x 4.0 cm, shape, location and cortical echogenicity.} &  
\begin{tabular}[t]{l}\footnotesize{(kidney, is, normal)}\\
\footnotesize{(right kidney, is, normal)}\\
\footnotesize{(right kidney, is normal in, size)}\\
\footnotesize{(kidney, is normal in, size)}\\

\end{tabular} \\
\hline 

\end{tabular}
}
\caption{Examples of triples extracted using OpenIE tool for given input sentences}
\label{tbl:openie}
\end{table}

Our approach combines rule-based, lexicon dictionary-based, and pattern-based techniques. The three modules of the IE task are (i) Lexical Analyser, (ii) Dependency Parser, and (iii) Semantic Analyser.
\\
\paragraph{\textbf{Lexical Analyser}}
There are three main tasks in the lexical analyser (i) Lexical semantic tagging, (ii) Noun phrase chunking and (iii) Medical lexicon dictionary creation. The input to the lexical analyser is a sentences from the corpus, and it outputs the sentence-wise syntactic and semantic features of each word or phrase in the sentence. Features include POS tag, lemma, supersense, and the root of a noun chunk.

\begin{figure}[h]
\centering
\includegraphics[width=0.8\columnwidth]{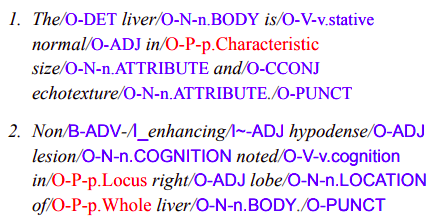}
\caption{Lexical-semantic supersense tagger tags all words with corresponding supersenses. We are interested in only supersenses of prepositions; hence those are highlighted in red. According to the context of sentence one and sentence two, tagger tagged preposition \textit{\color{darkgray}in} with different tags.}
\label{fig:ss_sample}
 \end{figure}

\subparagraph{\textbf{Lexical Semantic Supersense Tagger:}}
Prepositions play a role in forming relations between two entities. To extract the relation between two entities connected by preposition, the machine should understand the meaning of that preposition. As shown in \autoref{fig:ss_sample}, intuition of \textit{\color{darkgray}in} preposition in first sentence is \textbf{\small{characteristic}} and in second sentence is \textbf{\small{locus}}. Supersenses \citep{schneider2015hierarchy} helps to get the intuition of these prepositions. Supersenses disambiguate the lexical units by semantically classifying them. This \citep{liu2020lexical} paper deals with three sets of supersense labels: nominal,
verbal, and prepositional/possessive. Lexical Semantic Recognition(LSR)\footnote{\url{https://github.com/nelson-liu/lexical-semantic-recognition}} \citep{liu2020lexical} task effectively tags the supersense for each word in sentence.
\autoref{fig:ss_sample} shows the sentences that contain prepositions and their corresponding tags.

We have mapped preposition supersenses with corresponding logical relations. \autoref{tbl:ss} shows the examples of supersense classes and their corresponding logical relations. 
\begin{table}[htb]
\centering
\resizebox{0.9\columnwidth}{!}{%
\begin{tabular}{lc} 
\hline
\textbf{Supersense} & \textbf{Relation} \\
\hline
Locus & FoundIn \\

Gestalt & PartOf \\

PartPortion & PartOf \\
\hline
\end{tabular}
\begin{tabular}{lc} 
\hline
\textbf{Supersense} & \textbf{Relation} \\
\hline
Whole & PartOf \\
 
Manner & PropertyOf \\
Purpose & PropertyOf \\ 
\hline
\end{tabular}

}
\caption{The examples of supersenses and their corresponding logical relations.}
\label{tbl:ss}
\end{table}

\subparagraph{\textbf{Noun Phrase Chunker:}}
Noun phrase chunking is applied to get the candidate entity phrases. We use spacy APIs to get the noun phrases and corresponding root elements. For example, in sentence \textit{\color{darkgray}An area of increased echogenicity in the right lobe of liver, suggestive of focal fatty infiltration.} noun chunks are: \textit{\color{darkgray}an area}, \textit{\color{darkgray}increased echogenicity}, \textit{\color{darkgray}the right lobe}, \textit{\color{darkgray}liver}, and \textit{\color{darkgray}focal fatty infiltration}. And their corresponding root words are \textit{\color{darkgray}area}, \textit{\color{darkgray}echogenicity}, \textit{\color{darkgray}lobe}, \textit{\color{darkgray}liver} and \textit{\color{darkgray}infiltration} respectively.

\autoref{lex_table} shows the combined output of supersense tagger and noun phrase chunker.
\begin{table*}[htb]
\centering
\resizebox{\textwidth}{!}{%
\renewcommand{\arraystretch}{2}
\begin{tabular}{|p{0.18\textwidth}|p{0.18\textwidth}|p{0.1\textwidth}|p{0.18\textwidth}|p{0.18\textwidth}|p{0.18\textwidth}|} 
\hline
\textbf{Noun Phrases/Words} & \textbf{Token List} & \textbf{Root Token} & \textbf{POS Tags} & \textbf{Lemmas} & \textbf{Supersences}

 \\
\hline
Non-enhancing hypodense lesion & [Non, -, enhancing, hypodense, lesion] & lesion & [ADJ, ADJ, VERB, ADJ, NOUN] & [non, \-, enhance, hypodense, lesion] & [B-ADV, I\_, I~\-ADJ, O\-ADJ, COGNITION] \\
\hline

noted & [noted] & noted & [VERB] & [note] & [cognition] \\
\hline
in & [in] & in & [ADP] & [in] & [Locus] \\
\hline
right lobe & [right, lobe] & lobe & [ADJ, NOUN] & [right, lobe] & [O\-ADJ, LOCATION] \\
\hline 
of & [of] & of & [ADP] & [of] & [Whole] \\
\hline 
liver. & [liver, .] & liver & [NOUN, PUNCT] & [liver] & [BODY, O\-PUNCT ] \\
\hline 

\end{tabular}
}
\caption{Output of the lexical analyser for input \textit{\color{darkgray}Non-enhancing hypodense lesion noted in right lobe of liver.}}
\label{lex_table}
\end{table*}

\subparagraph{\textbf{Medical Lexicon Dictionary Creator:}}
Radiology reports contain large number of medical terms including hundreds of descriptive or modifier terms that are complex in nature, it includes abbreviations, synonyms and proper names. These all medical terms are not found in single medical glossaries. RadLex \citep{langlotz2006radlex} lexicon is used to create radiology dictionary. RadLex is a comprehensive glossary of radiology terms. Radlex lexicon contains long phrases e.g. \textit{\color{darkgray}fat homogeneous background echotexture}. Here \textit{\color{darkgray}fat}, \textit{\color{darkgray}homogeneous} and \textit{\color{darkgray}background} are the modifiers of \textit{\color{darkgray}echotexture}. In radiology reports individual entities such as \textit{\color{darkgray}homegeneous echotexture}, \textit{\color{darkgray}fat echotexture}, \textit{\color{darkgray}background echotexture} or only \textit{\color{darkgray}echotexture} appears frequently. Hence, using NLP techniques we have extracted radiological terms from RadLex lexicon at granular level. We have divided these long phrases into individual entities with meaning in the radiology context. Also there are some radiological terms which frequently appear in radiology reports but not present in RadLex lexicon e.g. \textit{\color{darkgray}echopattern}, \textit{\color{darkgray}corticomedullary differentiation}, \textit{etc.} Entities that are missing in the Radlex lexicon, we have added from the corpus. We use the same categories that we have defined as ontology classes.
\autoref{table:dict} shows the examples of entities and categories in our dictionary.
\begin{table}[htb]
\centering
\resizebox{\columnwidth}{!}{%
\begin{tabular}{lc} 
\hline
\textbf{Entity} & \textbf{Category} \\
\hline
lesion & observation \\

cirrhosis & pathologic-finding \\

hepatitis & inflammation \\

size & property \\
\hline
\end{tabular}
\begin{tabular}{lc} 
\hline
\textbf{Entity} & \textbf{Category} \\
\hline
small & size-modifier \\
 
left lobe & anatomy \\
 ankle fracture & injury \\
 chronic liver disease & disease\\
\hline
\end{tabular}

}
\caption{Examples of entities and their corresponding categories present in our radiology dictionary.}
\label{table:dict}
\end{table}

\begin{figure*}[h]
\centering

\includegraphics[width=0.8\textwidth]{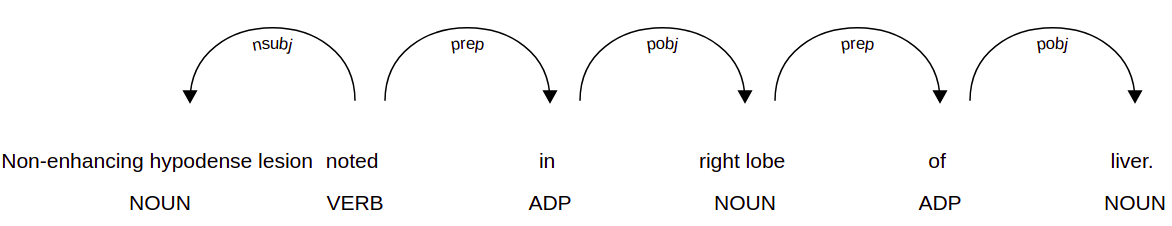}
\caption{Dependency tree of input sentence \textit{\color{darkgray}Non-enhancing hypodense lesion noted in right lobe of liver.}}
\label{fig:deptree}
\end{figure*}

\paragraph{\textbf{Dependency Parser}}
Dependency parser gives us direct or indirect linking between entity phrases. We have written rules over dependency tags and POS-tags by analyzing dependency parser to extract the relations between entities. Spacy\footnote{\url{https://spacy.io/api/dependencyparser}} APIs are used for dependency parsing. Dependencies are established between phrases instead of words. Before applying dependency parser we merge noun chunks into a single token. The dependency tree shown in the \autoref{fig:deptree}, \textit{\color{darkgray}Non-enhancing hypodense lesion} is linked with \textit{\color{darkgray}right lobe} and \textit{\color{darkgray}right lobe} is linked with \textit{\color{darkgray}liver}.

\paragraph{\textbf{Semantic Analyser}}
The previous modules' outputs help us write lexico-semantic rules and patterns to extract entities and relations from input sentences. In our case, entity extraction and relation extraction are very much dependent tasks. As dependencies are established between phrases, the dependency tree gives us the POS and dependency relation between noun phrases.

\subparagraph{\textbf{Dictionary-based Entity Extractor:} } Noun phrase chunker gives us noun phrases that are the candidate entity phrases. However, the entity may not always be a whole noun phrase. Hence to extract proper entities from noun phrases, we search the dictionary for matching entities. If a word or phrase matches with multiple dictionary entry through more than one text spans, we consider the longest text span as the matched entry for entity extraction. For example, in phrase \textit{\color{darkgray}right lobe} although individual terms \textit{\color{darkgray}right} and \textit{\color{darkgray}lobe} exist in our dictionaries, only the longest match, \textit{\color{darkgray}right lobe} is used for entity extraction.

\subparagraph{\textbf{Pattern-based Relation Extractor:}} Single noun phrase contains multiple entities. 
\begin{table*}[htb]
\centering
\resizebox{\textwidth}{!}{%
\renewcommand{\arraystretch}{2}
\begin{tabular}{|p{0.27\textwidth}|p{0.28\textwidth}|p{0.20\textwidth}|p{0.25\textwidth}|} 
\hline
\small{\textbf{Pattern}} & \small{\textbf{Triple Format}} & \small{\textbf{Example}} & \small{\textbf{Triples}}
 \\
\hline
\small{ADJ* NOUN/root} & \small{(ADJ, ModifierOf, NOUN/root)} & \small{simple clear cyst} & \small{(simple, ModifierOf, cyst)}\\ 
& & & \small{(clear, ModifierOf, cyst)}
\\
\hline 
\small{Anatomy Anatomy/root} & \small{(Anatomy/root, PartOf, Anatomy)} & \small{liver right lobe} & \small{(right lobe, PartOf, liver)}
\\
\hline

\small{Anatomy Finding/root} & \small{(Finding/root, FoundIn, Anatomy)} & \small{kidney calculus} & \small{(calculus, FoundIn, liver)}
\\
\hline
\small{Anatomy Observation/root} & \small{(Observation/root, FoundIn, Anatomy)} & \small{urinary bladder cyst} & \small{(cyst, FoundIn, urinary bladder)}
\\
\hline
\small{Modifier Observation/root} & \small{(Modifier, ModifierOf, Observation/root)} & \small{non-enhancing lesion} & \small{(non-enhancing, ModifierOf, lesion)}
\\
\hline
\end{tabular}
}
\caption{The list of some patterns and examples of triples extracted from noun phrases when patterns are applied to extract relations. /root represents the root entity of a noun phrase.}
\label{patn_table}
\end{table*}
Patterns are used to extract these entities. \autoref{patn_table} shows the dictionary of patterns. To extract entities we use look up based approach, that check the matching pattern in pattern dictionary for input noun phrase.  For example, \textit{\color{darkgray}non-enhancing} and \textit{\color{darkgray}hypodense} are the modifiers of \textit{\color{darkgray}lesion} in the noun phrase \textit{\color{darkgray}non-enhancing hypodense lesion}. As \textit{\color{darkgray}non-enhancing} present in the dictionary, it applies pattern \textbf{\small{Modifier Observation}} and extracts the triple (\textit{\color{darkgray}non-enhancing}, \textbf{\small{ModifierOf}}, \textit{\color{darkgray}lesion}). \textit{\color{darkgray}Hypodense} does not exist in our dictionary; hence, it applies the \textbf{\small{ADJ NOUN}} pattern and extracts triple (\textit{\color{darkgray}hypodense}, \textbf{\small{ModifierOf}}, \textit{\color{darkgray}lesion}). \autoref{lex_table} shows the list of some patterns and examples of triples extracted from noun phrases when patterns are applied to extract relations.
\\

\subparagraph{\textbf{Relation Extraction Using Preposition Supersenses:} } If two entities are connected with preposition then we consider its supersense to find the relation. For example, \textit{\color{darkgray}lesion in right lobe} here \textit{\color{darkgray}in} represents \textbf{\small{locus}} supersense and as shown in \autoref{tbl:ss}, \textbf{\small{locus}} is mapped to \textbf{\small{FoundIn}} relation. We add new triple ( \textit{\color{darkgray}lesion}, \textbf{\small{FoundIn}},  \textit{\color{darkgray}right lobe}).
\\
\\
\subparagraph{\textbf{Rule-based Relation Extractor:}} We discussed how to extract entities and the relation between the entities present in the single noun phrase. However, a relation exists between the entities present in two different noun phrases. The relation exists between root entities in candidate pair of noun phrases. The example shown in \autoref{fig:deptree}, there exist relation between \textit{\color{darkgray}lesion} and \textit{\color{darkgray}right lobe} but not between \textit{\color{darkgray}hypodense} and \textit{\color{darkgray}right lobe}.

We have written rules using the dependency tree to get the candidate pair of noun phrases that are related to one another. We write an algorithm to traverse the dependency tree from each leaf up to the root node to find the candidate pair of noun phrases. For example, the sentence given in \autoref{fig:deptree} there are two leaf nodes, \textit{\color{darkgray}Non-enhancing hypodense lesion} and \textit{\color{darkgray}liver}. Hence first it traverse from leaf node  \textit{\color{darkgray}Non-enhancing hypodense lesion} to root node \textit{\color{darkgray}noted} and then it traverse from \textit{\color{darkgray}liver} to root node \textit{\color{darkgray}noted}. We use different stacks to store important linguistic information like subjects, objects present in the current path, prepositions present in current path, \textit{etc.} 
We first find all the subjects present in path while traversing from leaf node to root node. We save that subjects globally with corresponding verb so that when we traverse different path from leaf to root then we can have access to all subjects that we have already traversed associated with verb. For example, we store \textit{\color{darkgray}lesion} (root entity of noun phrase \textit{\color{darkgray}Non-enhancing hypodense lesion}) as subject associated with verb \textit{\color{darkgray}noted}. Similarly we store all objects globally that occurs in the sentence associated with verb. For example, dependency tree shown in \autoref{fig:deptree} there exist path from leaf node \textit{\color{darkgray}liver} to root node \textit{\color{darkgray}noted}. In that path two objects are present \textit{\color{darkgray}liver} and \textit{\color{darkgray}right lobe}. When we process \textit{\color{darkgray}liver} it stores \textit{\color{darkgray}liver} in object stack and when it comes to \textit{\color{darkgray}right lobe} at that instance there is one preposition in preposition stack and one object in object stack. It pops \textit{\color{darkgray}of} from preposition stack and \textit{\color{darkgray}liver} from object stack. Supersense associated with preposition \textit{\color{darkgray}of} is \textbf{\small{Whole}} and this supersense is mapped with logical relation \textbf{\small{PartOf}}. Hence we define relation between \textit{\color{darkgray}right lobe} (root entity of \textit{\color{darkgray}right lobe} is \textit{\color{darkgray}right lobe}) and \textit{\color{darkgray}liver} is \textbf{\small{PartOf}}. Add new triple (\textit{\color{darkgray}right lobe}, \textbf{\small{PartOf}}, \textit{\color{darkgray}liver}) in triple set and push new object \textit{\color{darkgray}right lobe} in stack. Once it reaches to root node it adds new triples (subject, relation, object). If subjects and objects are not connected by any
preposition, then we use lexical-semantic patterns to extract the relation between them. Patterns such as if subject is Observation (e.g. \textit{\color{darkgray}lesion}) and object is Anatomy (e.g. \textit{\color{darkgray}right lobe}) then relation exist between them is ObservedIn. Hence we add new triple (\textit{\color{darkgray}lesion}, \textbf{\small{ObservedIn}}, \textit{\color{darkgray}right lobe}). List of patterns used to extract relations between two entities are listed in \autoref{tbl:SO_pattern}.
 \\
\begin{table*}[htb]
\centering
\resizebox{\textwidth}{!}{%
\renewcommand{\arraystretch}{2}
\begin{tabular}{|p{0.25\textwidth}|p{0.24\textwidth}|p{0.23\textwidth}|p{0.28\textwidth}|} 
\hline
\small{\textbf{Pattern (entity1-category, entity2-category)}} & \small{\textbf{Triple Format}} & \small{\textbf{Example (entity1, entity2)}} & \small{\textbf{Triples}}
 \\
\hline
\small{(Anatomy, Anatomy)} & \small{(entity1, PartOf, entity2)} & \small{(right lobe, liver)} & \small{(right lobe, PartOf, liver)}\\ 

\hline 
\small{(Property, Anatomy)} & \small{(entity1, PropertyOf, entity2)} & \small{(echotexture, pancreas)} & \small{(echotexture, PropertyOf, pancreas)}\\ 

\hline 
\small{(Finding, Anatomy)} & \small{(entity1, FoundIn, entity2)} & \small{(medical renal disease, kidney)} & \small{(medical renal disease, FoundIn, kidney)}\\ 

\hline 
\small{(Observation, Anatomy)} & \small{(entity1, ObservedIn, entity2)} & \small{(pseudo cyst, body)} & \small{(pseudo cyst, ObservedIn, body)}\\ 

\hline
\end{tabular}
}
\caption{The list of some patterns and examples of triples extracted when patterns are applied to extract relations between two entities.}
\label{tbl:SO_pattern}
\end{table*}

\subsubsection{Knowledge Graph Augmentation}
Preliminary KGs enhanced using triples extracted by IE module. Steps involved in KG augmentation are explained below:
\begin{itemize}
  \item \textbf{Step 1:} Triples are stored in the file against its input sentence. 
For example, sentence from corpus is, \textit{\color{darkgray}A lesion of increased echotexture in the right lobe of liver.} Triplets extracted corresponding to above sentence are, (\textit{\color{darkgray}increased, ModifierOf, echotexture}), (\textit{\color{darkgray}echotexture, PropertyOf, lesion}), (\textit{\color{darkgray}right lobe, PartOf, liver}), and (\textit{\color{darkgray}lesion, FoundIn, right lobe}).

\item \textbf{Step 2:} Construct dynamic KG for sentence triples.
\autoref{fig:DynamicKG} shows the dynamic KG constructed for sentence triples.

\begin{figure}[h]
\centering
\includegraphics[width=0.45\columnwidth]{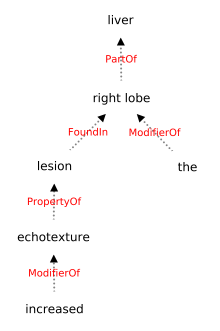}
\caption{Dynamic KG constructed for the triples of sentence \textit{\color{darkgray}\color{darkgray}A lesion of increased echotexture in the right lobe of liver.}}
\label{fig:DynamicKG}
\end{figure}

\item \textbf{Step 3:} Find its appropriate matched path in our already built preliminary (static) KG. 
\autoref{fig:StaticKG} shows the entities from dynamic KG path matched with static KG path.

\begin{figure}[h]
\centering
\includegraphics[width=0.6\columnwidth]{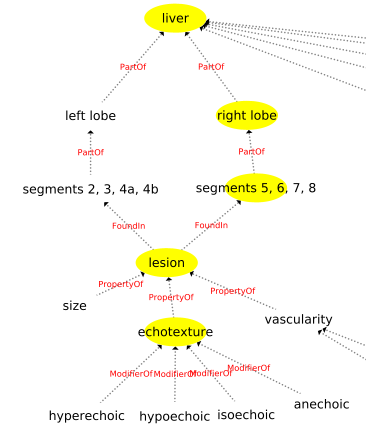}
\caption{The static KG at that instance. Yellow highlighted nodes show the entities from the dynamic KG path matched with the static KG path.}
\label{fig:StaticKG}
\end{figure}

\item \textbf{Step 4:} If a triple is missing in the static KG, then we add a new triple in the static KG. 
Here in above example triple (\textit{\color{darkgray}increased, ModifierOf, echotexture}) is missing in static KG. Hence, we will add this triple in static KG. \autoref{fig:UpdatedStaticKG} shows the updated static KG.

\begin{figure}[h]
\centering
\includegraphics[width=0.65\columnwidth]{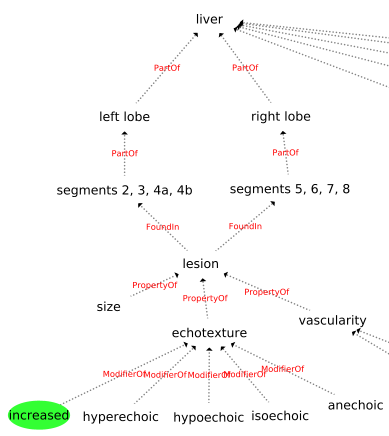}
\caption{Updated static KG after adding new triple. Newly added node highlighted in green.}
\label{fig:UpdatedStaticKG}
\end{figure}

\end{itemize}

This is how we update the static KG according to our dynamic KG triples. We repeat above steps for all sentences in our corpus.

\begin{figure*}[h]
\includegraphics[width=\textwidth]{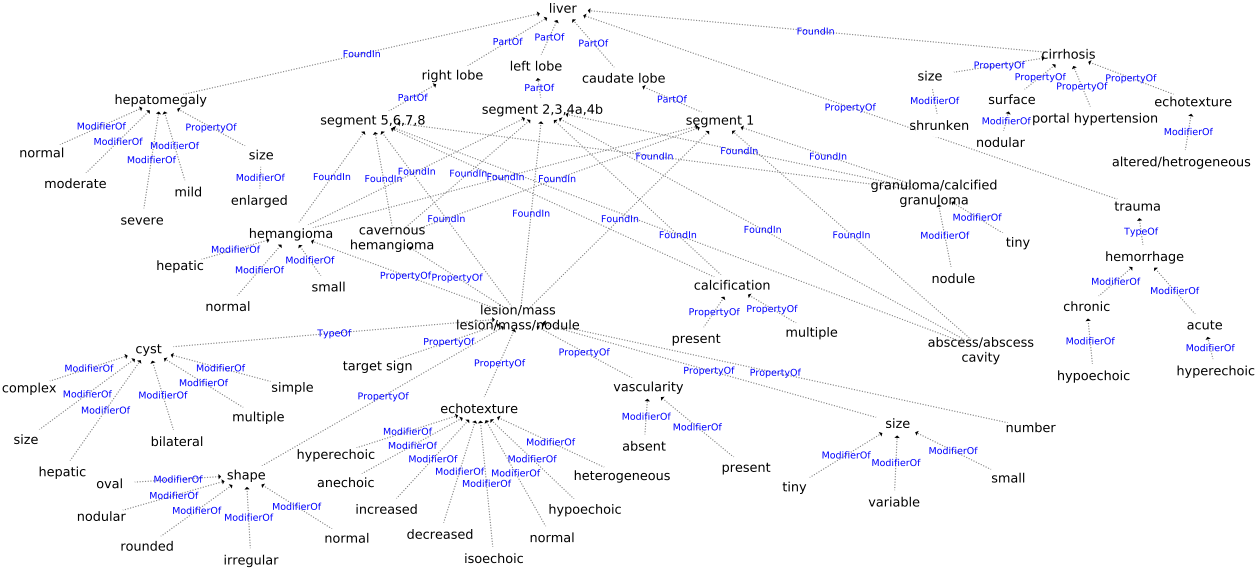}
\caption{The augmented KG of the Liver is completed by information extracted from the radiology report corpus. This figure shows a partial KG since it is large and can not represent it in limited page size.}
\label{fig:liveraugKG}
\end{figure*}

In a static KG, we have multiple instances of the same observations, same properties, and same modifiers.
For example, \textit{\color{darkgray}acute hepatitis} reveals \textit{\color{darkgray}decreased echogenicity} of the \textit{\color{darkgray}liver} and  \textit{\color{darkgray}chronic liver disease} reveals \textit{\color{darkgray}increased echogenicity} of the \textit{\color{darkgray}liver}. Here for both the findings \textit{\color{darkgray}echogenicity} is the related observation but their related descriptors/modifiers are not same. \textit{\color{darkgray}Decreased} is the \textit{\color{darkgray}echogenicity} modifier associated with \textit{\color{darkgray}acute hepatitis} and \textit{\color{darkgray}increased} is the \textit{\color{darkgray}echogenicity} modifier associated with \textit{\color{darkgray}chronic liver disease}. Therefore we have created different instances of observation with name \textit{\color{darkgray}echogenicity} for both the findings. Hence, we use a path from the dynamic KG to find the appropriate entity with identical names from the static KG.
In static KGs, we have arranged findings in such a way that its parents represent the anatomical location and its children represent the properties or states of organs related to that finding. \autoref{fig:liveraugKG} shows the augmented KG of the Liver. 

\subsection{Generating Pathological Description}
\begin{figure}[h]
\centering
\includegraphics[width=\columnwidth]{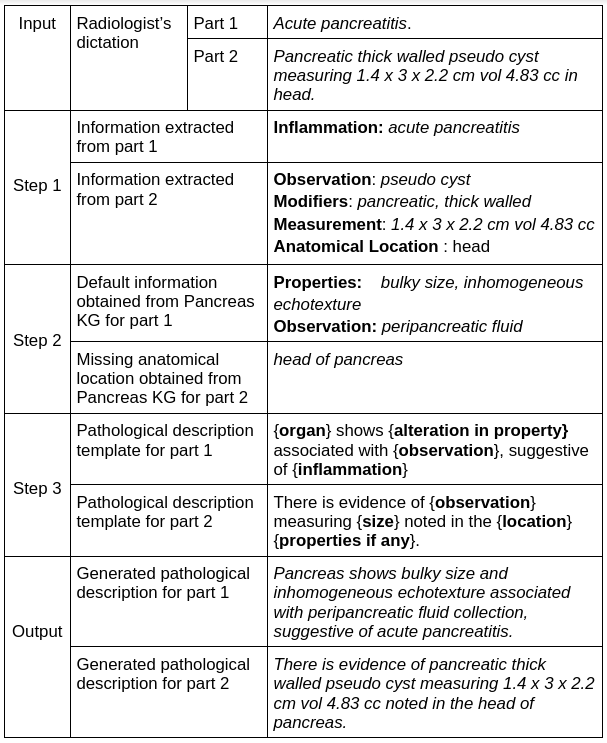}
\caption{Example of step-wise generated pathological description from radiologist's dictation.}
\label{fig:path_temp}
\end{figure}

To generate pathological descriptions, we first extract the essential information from dictation using the same IE module we built for the KG construction pipeline. We obtain the missing and default information from KGs and patient-specific information from dictation. Then fill this information in pathological description templates. We have defined pathological description templates based on the type of findings—for example, template for \textbf{inflammation, lesion, disease,} \textit{etc.} Table in the \autoref{fig:path_temp} shows the examples of generated pathological description from radiologist's dictation.

\subsection{Generating Patient Specific Report}
One of the crucial tasks of the project is the replacement of normal descriptions with the generated pathological descriptions into the normal report template.
Default normal report templates are provided by the radiologists. We provide an appropriate normal report template to the system, and the system replace generated pathological descriptions with the corresponding location in the normal report template.
\\
\subsubsection{Parallel Corpus}
At first, we create a parallel corpus for the radiologist’s dictation and the corresponding normal sentences in a normal report template (referred to as \textbf{\small{normal description}}).
\begin{table}[htb]
\centering
\resizebox{\columnwidth}{!}{%
\renewcommand{\arraystretch}{2}
\begin{tabular}{l p{0.4\columnwidth} p{0.6\columnwidth}} 
\hline
 & \textbf{\small{Radiologist's dictation}} & \textbf{\small{Normal description}} \\
\hline
i. & \small{Grade 1 fatty liver} & \small{Liver is normal in size and echotexture.} \\

ii. & \small{Acute pancreatitis} & \small{Pancreas is normal in size and echotexture.} \\

iii. & \small{Chronic pancreatitis} & \small{Pancreas is normal in size and echotexture.} \\

iv. & \small{Pancreatic thick walled pseudo cyst in head.} & \small{No evidence of focal or diffuse lesion is seen.} \\

v. & \small{Pancreatic thick walled pseudo cyst in tail.} & \small{No evidence of focal or diffuse lesion is seen.}\\
\hline
\end{tabular}
}
\caption{Samples from parallel corpus of radiologist's dictation vs. corresponding sentences from normal report template}
\label{table:parallelCorpus}
\end{table} 

Example shown in the \autoref{fig:path_temp}, generated pathological description for dictation \textit{\color{darkgray}acute pancreatitis}, \textit{\color{darkgray}size} and \textit{\color{darkgray}echotexture} is altered in pathological description, we replace the normal description about the \textit{\color{darkgray}size} and \textit{\color{darkgray}echotexture} of the Pancreas with generated pathological description in the normal report template. 
\\
\subsubsection{Workflow}
We have implemented a look-up-based approach to achieve this task. The basic approach says that for a radiologist's dictation, search for the corresponding normal description in a parallel corpus. Then find the location of that normal description in the normal report template and replace the whole sentence with the generated pathological description.

\begin{itemize}
    \item Input:
    \begin{enumerate}
        \item Radiologist's dictation
        \item Generated pathological description
        \item Parallel corpus of radiologist's dictation vs. normal descriptions
        \item Normal report template
    \end{enumerate}
    \item Output: Radiology report with patient-specific findings.
\end{itemize}

\begin{figure*}[h]
\centering
\includegraphics[width=\textwidth]{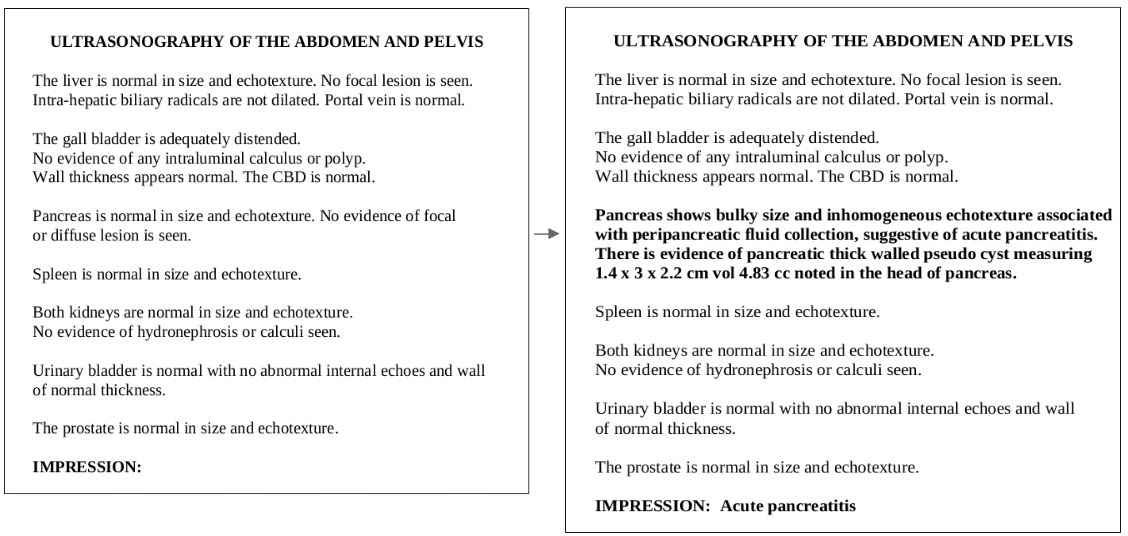}
\caption{Left hand side shows normal report template of ultrasonography of the Abdomen and Pelvis, and the right hand side shows a patient-specific report of ultrasonography of the Abdomen and Pelvis.}
\label{fig:normalReport}
\end{figure*}

\begin{itemize}
  \item \textbf{Step 1:} Perform a look up into a parallel corpus to find similar radiologist's dictation with input radiologist's dictation. We use NLTK BLEU score for matching the dictations. Get the corresponding normal description from corpus. For example, for input dictation \textit{\color{darkgray}Acute pancreatitis} we found match \textit{\color{darkgray}Acute pancreatitis} in dictionary. Similarly for input dictation \textit{\color{darkgray}Pancreatic thick walled pseudo cyst measuring 1.4} x \textit{\color{darkgray}3} x \textit{\color{darkgray}2.2 cm vol 4.83 cc in head.} we found corresponding matched dictation \textit{\color{darkgray}Pancreatic thick walled pseudo cyst in head.}
  
As shown in \autoref{table:parallelCorpus}, pathological description corresponding to matched dictations are \textit{\color{darkgray}Pancreas normal in size and echotexture} and \textit{\color{darkgray}No evidence of focal or diffuse lesion is seen.}
  \item \textbf{Step 2:} In given normal report template find the appropriate normal sentence to replace with generated pathological description. We find sentences similar to normal description found in step 1 to replace in template (left side of the \autoref{fig:normalReport} shows normal report template). 

  \item \textbf{Step 3:} Replace matched normal sentences in the template with the corresponding generated pathological description. \autoref{table:replaceBy} shows the matched sentences in template to replace with generated pathological descriptions.
    \begin{table}[htb]
        \resizebox{\columnwidth}{!}{%
            \renewcommand{\arraystretch}{2}
            \begin{tabular}{l p{0.4\columnwidth} p{0.6\columnwidth}} 
                \hline
                & \small{\textbf{Sentence to replace}} & \small{\textbf{Replace by}} \\
                \hline
                i. & \small{Pancreas is normal in size and echotexture.} & \small{Pancreas shows bulky size and inhomogeneous echotexture associated with peripancreatic fluid collection, suggestive of acute pancreatitis.} \\
                
                ii. & No evidence of focal or diffuse lesion is seen. & \small{There is evidence of pancreatic thick walled pseudo cyst measuring 1.4 x 3 x 2.2 cm vol 4.83 cc noted in the head of pancreas.} \\
                \hline
            \end{tabular}
            }
            \caption{Column 1 shows the sentences in normal report template to replace by corresponding generated pathological descriptions shown in column 2. }
            \label{table:replaceBy}
    \end{table}
  
\end{itemize}

Left hand side of the \autoref{fig:normalReport} shows normal report template and right hand side of the \autoref{fig:normalReport} shows the patient-specific report after replacing normal sentences with generated pathological descriptions.
One of the risks of our system is if the information does not exist in KG, then it only considers patient-specific information but not a static one.

\section{Evaluation}
\subsection{Evaluation of Information Extraction Task}
To evaluate our IE system, we chose method that calculates the precision, recall, and F-measure by manually comparing the machine annotations against the gold standard.
The evaluation method of precision and recall is explained by \cite{fader2011identifying} and \cite{etzioni2011open} for information retrieval tasks. \cite{etzioni2011open} and \cite{fader2011identifying} defined precision as the fraction of returned extracted correct triples and recall as the fraction of correct triples in the total corpus. For each extracted triple in a sentence, we manually check whether it is correct or not against the gold standard triples of the corresponding sentence. We calculate precision and recall for each sentence, then calculate average precision and recall. We calculate F1-Score on average precision and average recall. \autoref{tbl:ie_score} shows the results of our IE system and OpenIE system. 

\begin{table}[htb]
\centering
\resizebox{0.7\columnwidth}{!}{%
\begin{tabular}{l|ccc} 
\hline
\textbf{} & \textbf{Precision} & \textbf{Recall} & \textbf{F1-Score} \\
\hline
\textbf{Our System} & 0.93 & 0.92 & 0.92 \\

\textbf{OpenIE} & 0.57 & 0.60 & 0.58 \\

\hline
\end{tabular}

}
\caption{First row in the table shows the precision, recall and F-Score for triples extracted by our system vs. gold standard triples. Second row in the table shows the precision, recall and F-Score for entity pair from triples extracted by OpenIE tool vs. entity pairs in gold standard triples. }
\label{tbl:ie_score}
\end{table}

For OpenIE annotations, we considered only entity pairs from triple but no relation. Relations in our gold standard annotations are the logical relations that we have already defined in our IE pipeline. Relations given by OpenIE are verb-based relations from the sentence itself. Hence, we do not consider relations for calculating precision, recall, and F1-Score of OpenIE annotations vs. gold standard triples.

\subsection{Evaluation of Knowledge Graph}
The efficacy of KGs is tested after generating pathological descriptions using KG and their corresponding gold standard pathological descriptions generated by radiologists. We calculate BLEU\footnote{\url{https://www.nltk.org/\_modules/nltk/translate/bleu\_score.html}}, and ROUGE\footnote{\url{https://pypi.org/project/rouge-score/}} score metrics. To calculate the evaluation metrics, we have used 170 samples of Pancreas pathological descriptions. Table \autoref{tbl:blue} shows the BLEU score and \autoref{tbl:rouge} shows the ROUGE scores.

\begin{table}[htb]
\centering
\resizebox{0.55\columnwidth}{!}{%
\begin{tabular}{l|c} 
\hline
\textbf{} & \textbf{BLEU score}\\
\hline
\textbf{1-gram} & 0.5839 \\

\textbf{2-gram} & 0.4011 \\

\textbf{3-gram} & 0.3206 \\

\textbf{4-gram} & 0.2843\\
\textbf{cumulative 4-gram} & 0.3613\\
\hline
\end{tabular}

}
\caption{BLEU score of pathological descriptions generated by our system vs. gold standard pathological descriptions.}
\label{tbl:blue}
\end{table}
\begin{table}[htb]
\centering
\resizebox{0.7\columnwidth}{!}{%
\begin{tabular}{c|c c c} 
\hline
& \textbf{Precision} & \textbf{Recall} & \textbf{F-Measure}\\ 

\hline
\textbf{ROUGE-1} & 0.7450 & 0.6667 & 0.6937 \\

\textbf{ROUGE-2} & 0.4848 & 0.4395 & 0.4572 \\

\textbf{ROUGE-3} & 0.3705 & 0.3374 & 0.3508  \\

\textbf{ROUGE-4} & 0.2938 & 0.2673 & 0.2779 \\

\textbf{ROUGE-L} & 0.7034 & 0.6311 & 0.6573  \\
\hline
\end{tabular}

}
\caption{ROUGE score of pathological descriptions generated by our system vs. gold standard pathological descriptions.}
\label{tbl:rouge}
\end{table}

Two drawbacks of n-gram-based metrics are: i) such methods often fail to match
paraphrases robustly, and ii) n-gram models penalize semantically-critical ordering changes since it fail to capture distant dependencies \citep{zhang2019bertscore}. BERT-based
embeddings for sentence similarity addressed the above two pitfalls.
Hence, we have evaluated the cosine similarity of system-generated
pathological descriptions vs. gold standard pathological descriptions
using BioBERT \citep{lee2020biobert} sentence embeddings. The similarity score is \textbf{0.97}.

Also, system-generated pathological descriptions are evaluated by radiologists manually. In \textbf{80-85} percent of cases, generated pathological
descriptions are correct.

\section{Conclusion and Future Work}
Our approach has established a systematic process to construct organ-wise KGs of radiology concepts from free-text radiology reports for different organs. Also, we have introduced a combined approach based on linguistic rules, dictionaries, patterns, and preposition supersenses to extract radiological entities and their relations. We have constructed generic IE pipeline that can be used for other radiology scan reports like CT, MRI, X-Ray, \textit{etc.}
Domain experts evaluated constructed KGs of static information that is high-quality KGs. The KGs are stored in standard RDF format; hence that can be applied to various medical domain applications. Currently, we are using these constructed KGs to generate structured patient-specific radiology reports. Using domain-specific KG in downstream NLP applications will eliminate the annotated data requirement.

The limitation of our work is that to generate patient-specific reports, we use standard predefined pathological description templates and standard normal report templates. These are not customized according to the radiologists' reporting style.

We plan to customize generated reports according to the radiologists' reporting style. Also, we plan to combine all KGs of different organs in a single KG in the future. That would be a single knowledge base for all organs. A single KG will help if a single narrated sentence includes a pathological description of multiple organs. We have constructed KGs for Ultrasound scan procedure. We plan to construct KGs for other scan procedures like CT, MRI, X-Ray, \textit{etc.}, using above mentioned KG construction module.

\bibliography{custom,anthology}
\bibliographystyle{acl_natbib}
\appendix

\section{Constructed Knowledge Graphs Examples}

\begin{figure*}[h]
\centering
\makebox[\textwidth]{\includegraphics[width=0.9\paperwidth]{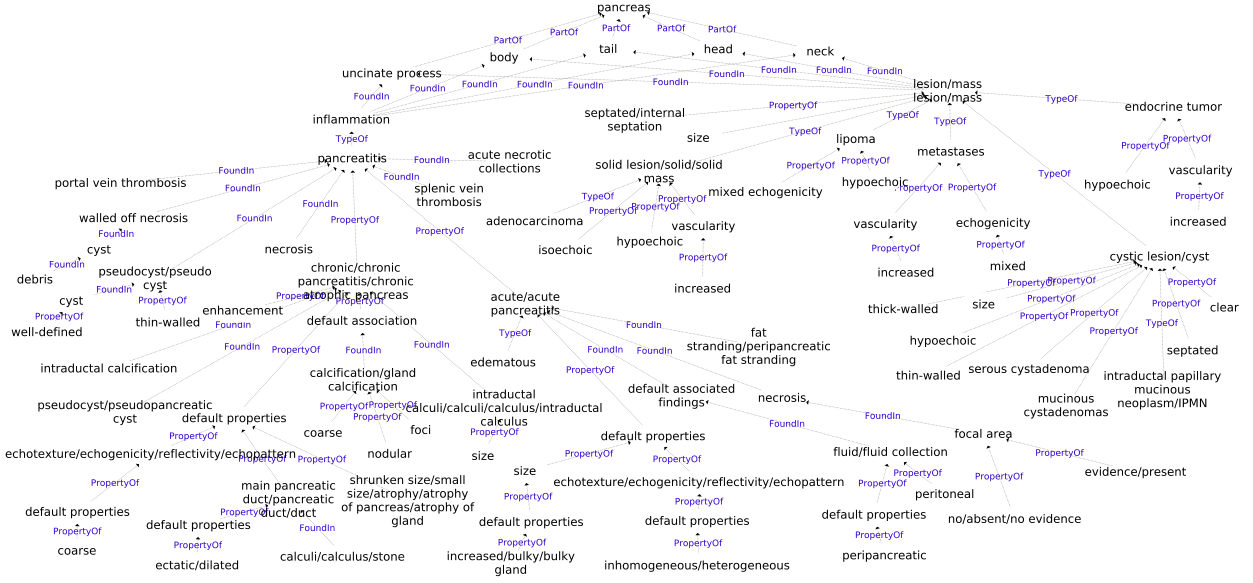}}
\caption{Knowledge graph of the Pancreas constructed using knowledge graph construction module.}
\label{fig:pancreas_KG}
\end{figure*}

\begin{figure*}[h]
\centering
\makebox[\textwidth]{\includegraphics[width=0.9\paperwidth]{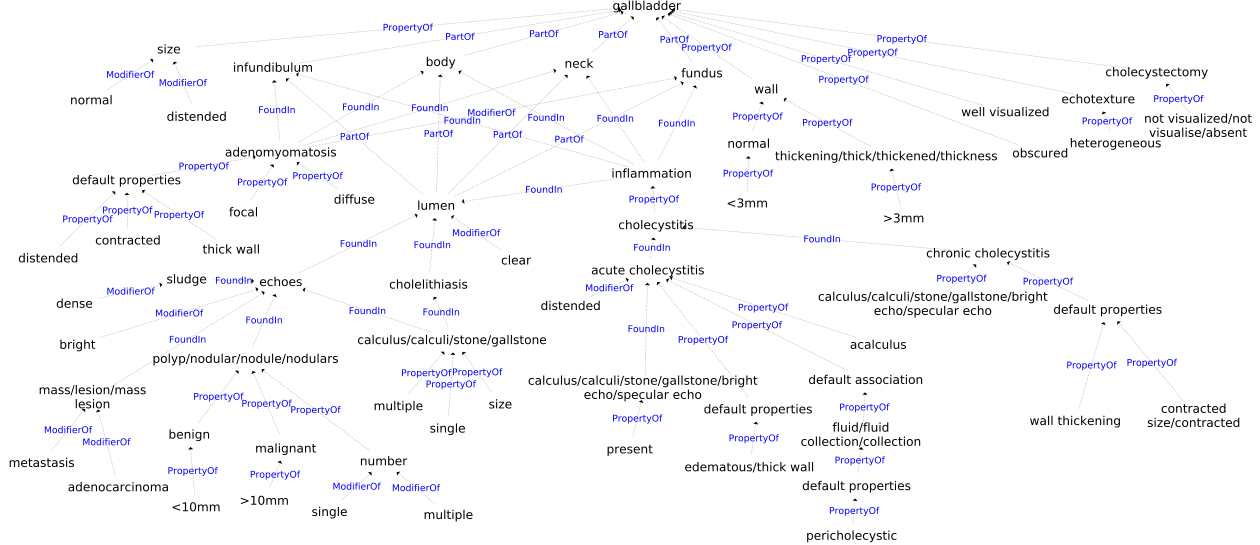}}
\caption{Knowledge graph of the Gallbladder constructed using knowledge graph construction module.}
\label{fig:gb_kg}
\end{figure*}

\end{document}